\documentclass{isprs} 
\usepackage{subfigure}
\usepackage{setspace}
\usepackage{geometry} 
\usepackage{epstopdf}
\usepackage{tikz}
\usepackage{amssymb}
\usepackage{amsmath}
\usepackage{amsfonts}
\usepackage[labelsep=period]{caption}  
\usepackage[british]{babel} 
\usepackage[hang]{footmisc}
\usepackage{multirow}
\usepackage{booktabs}
\usepackage{siunitx}
\usepackage{makecell}
\usepackage{hyperref}
\usepackage{booktabs}
\usepackage{tabularx}
\usepackage{multirow}
\usepackage{makecell}
\usepackage{siunitx}
\usepackage{booktabs}
\usepackage{tabularx}
\usepackage{siunitx}
\usepackage{array} 

\hypersetup{
    colorlinks = true,
    linkcolor = blue,
    citecolor = blue,
    urlcolor  = blue
}

\begin{document}

\title{Extrusion Segmentation Strategy to improve CAD Reconstruction from Point Cloud
}
\date{}
\author{
 Said Harb, Mehdi Maboudi, Markus Gerke}
\address{
	Institute of Geodesy and Photogrammetry, Technische Universit\"at Braunschweig, Germany \\{(s.harb, m.maboudi, m.gerke)}@tu-braunschweig.de
}



\abstract{
Computer-Aided Design is ubiquitous in today’s world, as almost every manufactured object begins as a digital model across industries. At the same time, advances in 3D sensing have made point clouds a dominant form of raw 3D data. Recovering the CAD model of a physical object from its point cloud scan has two major applications: reverse engineering, where physical or hand-crafted prototypes need to be reconstructed automatically as editable digital models, and quality control, where recovering the CAD description of a manufactured object helps quantify and understand deviations introduced during the production process. Thus, converting unordered point clouds into structured CAD models is increasingly important for modern applications. Deep learning has enabled major progress in computer vision for both 2D and 3D data, and new datasets facilitate data-driven CAD reconstruction. Building on this foundation, we develop an end-to-end model that reconstructs CAD models from point clouds and introduce a segmentation approach that decomposes them into individual extrusions. These partial shapes increase data diversity, improving the generalization and robustness of deep learning models. Our strategy thereby provides a simple, yet effective way to increase reconstruction performance of deep learning models.
}

\keywords{Point Cloud, CAD, 3D Reconstruction, Deep Learning, Segmentation}

\maketitle

\sloppy


\section{Introduction}
From civil engineering, aerospace, automotive, architectural design to manufacturing~\cite{Wu-2021_ICCV}, Computer Aided Design (CAD) models help engineers to design, analyze and collaborate on systems, before they are physically built. At the same time, point clouds have become one of the most prevalent forms of three-dimensional (3D) measurement data, largely due to advances in 3D sensing technologies~\cite{huang-2021_ICCV}. Point clouds are unordered sets of 3D points, and unlike CAD models, they lack connectivity and semantic information~\cite{farshian-2023_IEEE}. Therefore, there has been interest in converting unstructured point clouds into structured and concise CAD representations. We refer to this task as reverse modeling, emphasizing that the physical object already exists, and the goal is to reconstruct the original modeling process that produced it. The importance of CAD models is further underscored by the projected growth of the 3D CAD software market, expected to rise from 13.40 billion USD in 2025 to 24.23 billion USD by 2034~\cite{sun-2025}. Another key application is quality control in a post-processing step: if geometric parameters can be derived from a produced object, the possible sources of misalignment and errors can be directly identified. As a consequence, tools or manufacturing workflows can be systematically optimized.

\begin{figure}[b]
    \centering
    \includegraphics[width=1\linewidth]{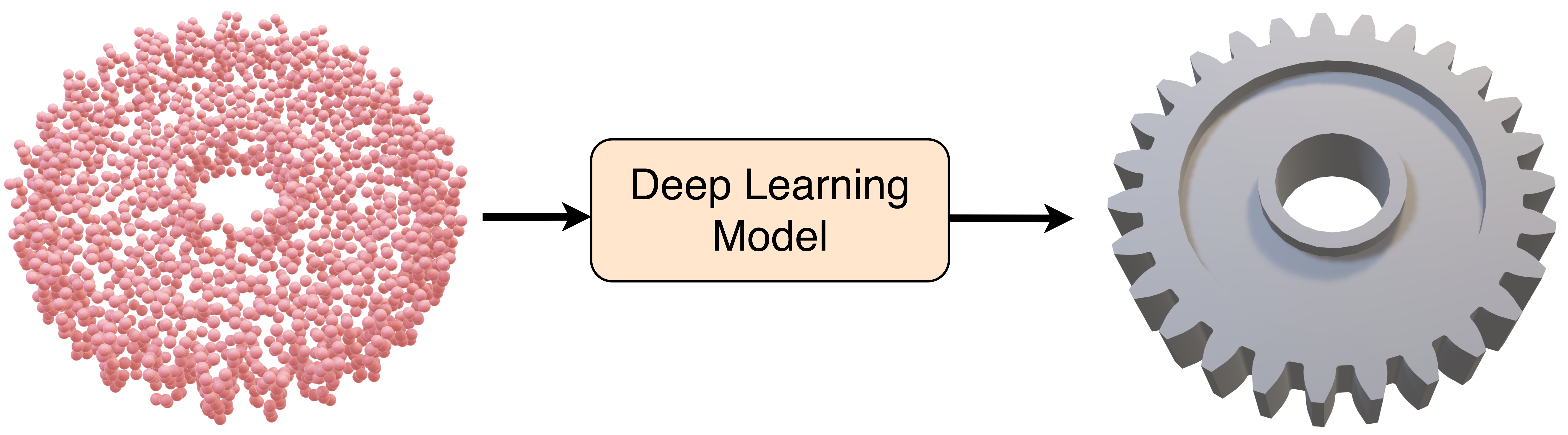}
    \caption{CAD models are reconstructed from input point clouds.}
    \label{fig:intro}
\end{figure}

Prior to the rise of Deep Learning (DL) methods, geometric algorithms such as Poisson surface reconstruction, ball-pivoting, or Delaunay triangulation were the dominant approaches for reconstructing geometric models from input point clouds. These methods, however, struggle to scale to diverse and detailed datasets and often fail to generalize to arbitrary data~\cite{farshian-2023_IEEE}. DL has enabled remarkable breakthroughs in Computer Vision, initially for two-dimensional (2D) data and, more recently in 3D domains as well. In contrast to traditional approaches such as Random Sample Consensus (RANSAC), which rely on manual assumptions about the data \cite{li-2018}, DL models can learn directly from the training data. Furthermore, typical challenges with point cloud data, such as outliers, inhomogeneous density, occlusions, or noise, can be incorporated into the training data~\cite{pc2model}, thereby improving the generalization ability of DL models. CAD modeling at the user level is a sequential, feature-based process that constructs shapes from geometric primitives (e.g., spheres, cubes, cylinders) using operations such as extrusion and Boolean operations~\cite{sharma-2017}. Each operation adds a feature to the model, forming a “design history” of sketch-and-extrude steps that convert 2D sketches into 3D geometry. To enable DL models to learn such CAD sequences, large-scale datasets capturing the construction history of CAD models are required. In this context, Wu et al.~\cite{Wu-2021_ICCV} introduced the DeepCAD dataset, a large sequence-based CAD dataset that facilitates DL-driven CAD reconstruction. It comprises 178{,}238 models represented in sequence format. 

The objective of this work is to develop an end-to-end pipeline that reconstructs CAD models in the form of CAD sequences from point clouds of object surfaces, depicted in Figure~\ref{fig:intro}. The proposed pipeline shows high-fidelity reconstruction for non-complex objects, especially primitive shapes such as cubes and hollow or solid cylinders, as well as assemblies of these forms. Yet reconstruction quality degrades on more complex shapes, which tend to integrate multiple features and exhibit more intricate geometric structures. To address this, we introduce an extrusion-based segmentation strategy that decomposes CAD models into their constituent extrusions, and we show that incorporating these partial shapes into the training data improves the generalization and robustness of DL models. This way, also more complex objects can be processed. These insights are potentially valuable for a variety of application areas, including robotics, Virtual/Augmented Reality (VR/AR) environments, additive manufacturing, simulation with real-world objects, and digital heritage preservation~\cite{uy-2021}.

Section~\ref{sec:related_work} provides the theoretical background and reviews related work relevant to this research. Section~\ref{sec:methodology} then describes the methodology employed in this study. The experimental setup used to train the deep learning model is presented in Section~\ref{sec:exp_setup}. In Section~\ref{sec:results}, the reconstruction quality is evaluated both qualitatively and quantitatively. Finally, Section~\ref{sec:conclusion} concludes the paper by summarizing the main findings, discussing practical use cases, and outlining directions for future work.

\section{Related Work} 
\label{sec:related_work}
This section begins by outlining the surface reconstruction problem and its inherent challenges. It then discusses the various surface representations that can be derived from point clouds, followed by a review of current research in this area, with particular emphasis on point cloud to CAD reconstruction and deep learning–based approaches.

Unlike structured or regular data representations, point clouds do not contain explicit connectivity information between points~\cite{farshian-2023_IEEE}. Real-world point clouds are typically acquired using techniques such as Light Detection and Ranging (LiDAR), photogrammetry, or multi-view imaging systems, among other methods~\cite{isprs-archives-XLIII-B1-2021-85-2021}. Alternatively, point cloud acquisition can be simulated in virtual environments such as Helios++, which allows modeling noise, beam divergence, and other measurement inaccuracies commonly encountered in real-world scans \cite{pc2model}. Furthermore, point clouds can be sampled directly from the surfaces of existing digital models in a controlled and reproducible manner.

\begin{figure}[b]
    \centering
    \includegraphics[width=\linewidth]{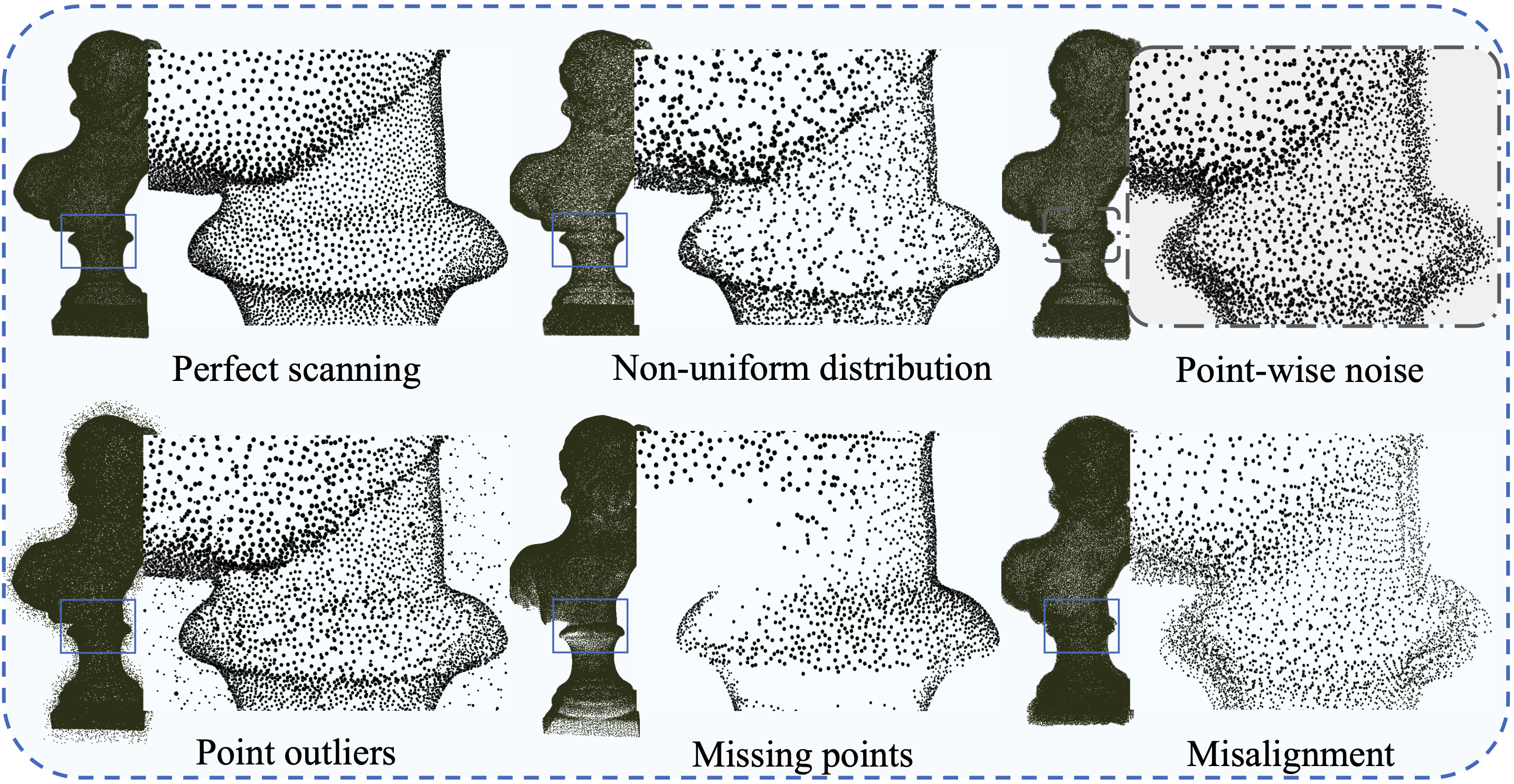}
\caption{Challenges in laser scanning. Adapted from~\protect\cite{huang-2024}.}
    \label{fig:scan_challenges}
\end{figure}

\begin{figure}[t]
    \centering
    \includegraphics[width=1\linewidth]{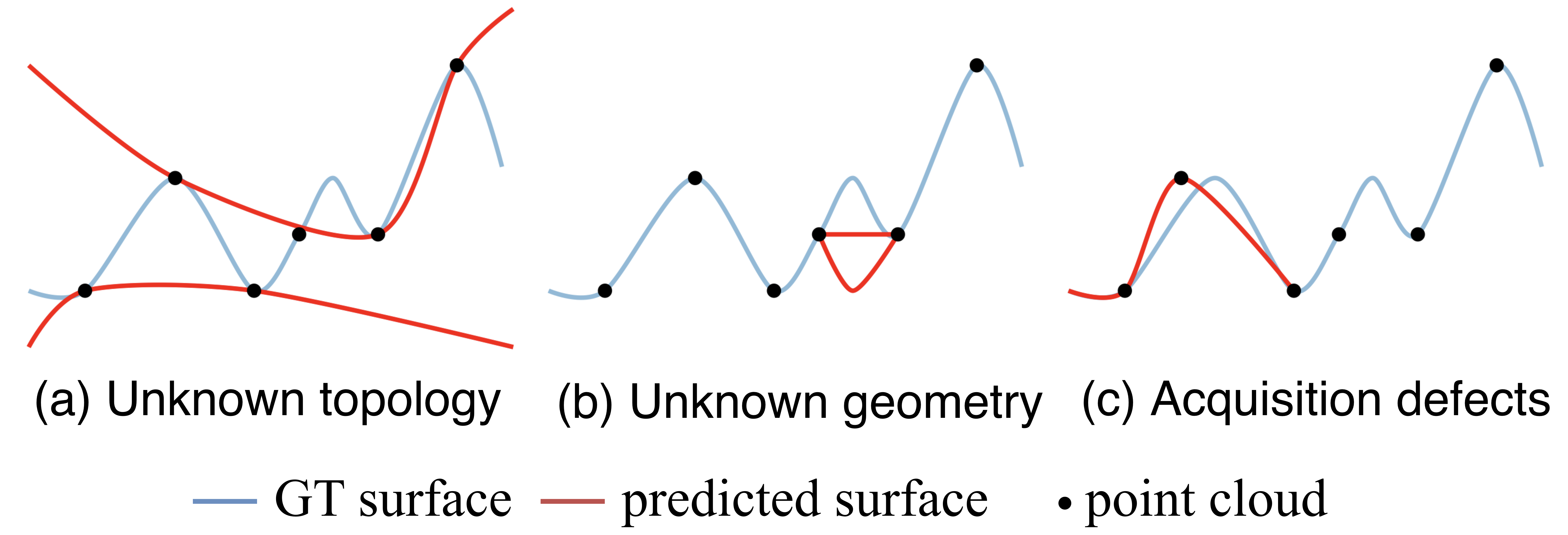}
    \caption{Surface reconstruction challenges. Adapted from \protect\cite{sulzer-2024}.}
\label{fig:reconstruction_challenges}
\end{figure}

\begin{figure}[b]
    \centering
    \includegraphics[width=1\linewidth]{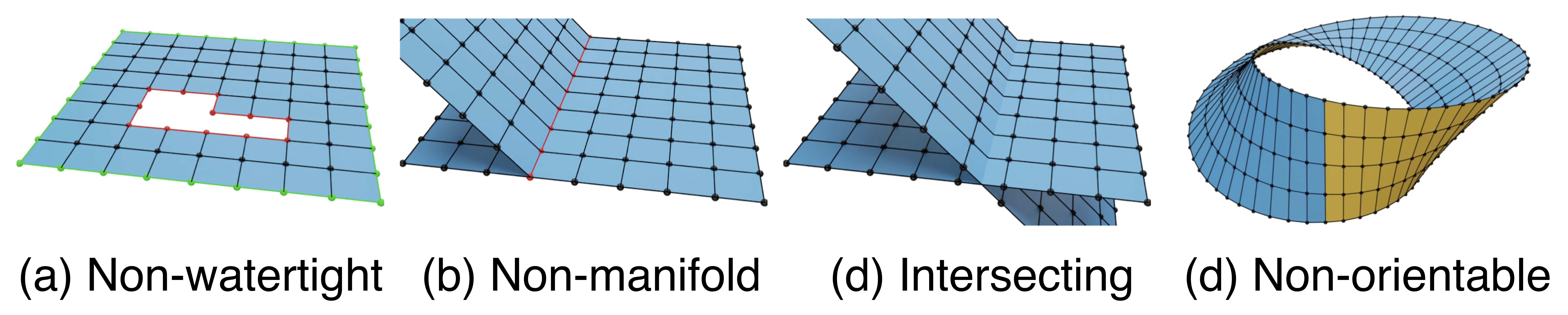}
    \caption{Examples of undesired properties in reconstructed surfaces: (a) a hole in the center (red), (b) an edge connects to three faces, (c) planes intersect, (d) no unique inside and outside orientation. Adapted from~\protect\cite{sulzer-2024}.}
    \label{fig:mesh_props}
\end{figure}

In practice, point cloud data are often imperfect. Typical challenges, illustrated in Figure~\ref{fig:scan_challenges}, include non-uniform point distributions, measurement noise, outliers, and missing data. Furthermore, when multiple scanners or viewpoints are used and point clouds from different perspectives are merged, misalignment between individual scans may occur~\cite{huang-2024}. From a surface reconstruction perspective, the problem is further complicated by the unknown geometry and topology of the underlying object. A single point set may correspond to multiple geometrically distinct surfaces, making surface reconstruction an inherently ill-posed problem. Given a point cloud, there is no unique solution for the surface it represents, instead, infinitely many plausible reconstructions may exist, as illustrated in Figure~\ref{fig:reconstruction_challenges}. Despite this ambiguity, a set of desirable properties can be defined that a reconstructed surface should ideally satisfy. First, the surface should be watertight, meaning that it contains no holes or gaps and fully encloses a finite volume. Second, the surface should be 2-manifold, implying that it locally resembles a two-dimensional surface. In practical terms, this requires that no edge is shared by more than two faces and that the faces incident to each vertex are arranged consistently without self-intersections, as shown in Figure~\ref{fig:mesh_props}. Additionally, surface components that are not directly connected should not intersect. Finally, the surface should be orientable, allowing for a consistent definition of an interior and exterior across the entire mesh. These properties are essential for ensuring that reconstructed surfaces are suitable for downstream applications such as physical simulation, visualization, and mesh editing \cite{sulzer-2024}.

\begin{figure}[t]
    \centering
    \includegraphics[width=1\linewidth]{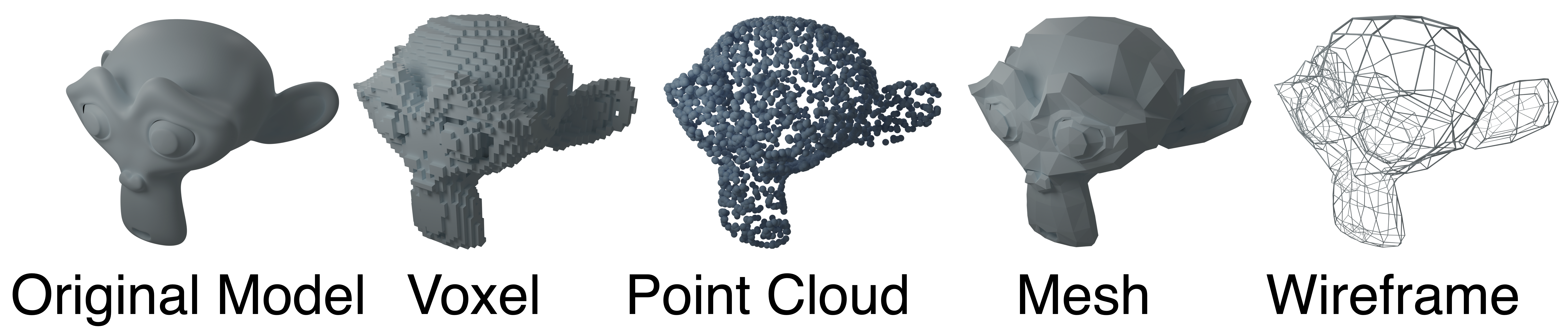}
    \caption{Different representations of a 3D model.}
    \label{fig:repre}
\end{figure}

To address the surface reconstruction problem for point clouds, researchers transform point clouds into various representations. One of the earliest and simplest approaches is the voxel representation, which partitions the 3D metric space into a regular grid of cuboidal cells, or voxels. This representation is compatible with conventional 3D deep learning techniques, such as Convolutional Neural Networks (CNNs). However, voxel grids are limited in resolution due to the discretization of continuous space, and as the grid resolution increases, the fraction of voxels containing meaningful data decreases~\cite{farshian-2023_IEEE}. Another widely used representation is the mesh, which explicitly models connectivity between points. A mesh is an irregular data structure composed of vertices, edges, and faces, allowing for accurate modeling of both geometry and topology. Mesh-based approaches can be divided into two main categories: patch-based and deformable-template-based methods. These representations have the advantage of directly generating a digital surface model of the predicted object. Nonetheless, challenges remain: patch-based methods often struggle to produce watertight meshes, while deformable-template methods require an initial mesh to deform, which may not adequately represent all inputs~\cite{farshian-2023_IEEE,sulzer-2024}. Instead of explicitly modeling surfaces, they can also be represented implicitly using neural representations. Here, a neural network acts as a universal function approximator for the 3D object. Implicit representations commonly employ occupancy functions or signed distance functions (SDFs). Occupancy functions predict whether a point in 3D space lies inside the object, while SDFs assign each point a signed distance to the object’s boundary, indicating whether it is inside or outside. Implicit representations have the advantage of being continuous and theoretically capable of infinite resolution, depending on the capacity of the neural network. However, post-processing algorithms are still required to extract an explicit surface from the network output~\cite{farshian-2023_IEEE}. Some of the representations are shown in Figure~\ref{fig:repre}.

CAD is ubiquitous in modern manufacturing and serves as the standard format for object design across industries~\cite{Wu-2021_ICCV}. In contrast to representations such as voxels, meshes, or point clouds, CAD models can be directly manipulated using CAD software, providing parametric control and precise geometric editing, capabilities that point clouds inherently lack~\cite{li-2018}. Moreover, CAD models satisfy the requirements for reconstructed surfaces illustrated in Figure~\ref{fig:mesh_props}. Reconstructing CAD models from scanned data is an important research problem, particularly when the original CAD model is unavailable~\cite{uy-2021}. This task is also relevant for reverse engineering, design exploration, and the generation of models from images or technical drawings~\cite{jayaraman-2023}. However, training computational agents to replicate the skills of professional CAD designers remains highly challenging~\cite{xu-2022}. Three main research directions have emerged in this domain: Constructive Solid Geometry (CSG) tree reconstruction, primitive fitting, and CAD sequence reconstruction. CSG tree reconstruction involves combining parametric primitives (e.g., cuboids, spheres, and cones) using Boolean operations to form a hierarchical shape representation~\cite{xu-2022}. In primitive fitting approaches, geometric primitives are directly fitted to input point clouds, enabling partial or global approximations of the target shape~\cite{li-2018}. CAD sequence reconstruction aims to recover the sequence of sketch-and-extrude operations that a designer would apply, where a closed 2D sketch is typically created and subsequently extruded into 3D geometry~\cite{uy-2021}.

One of the earliest works on learning CSG trees is CSGNet by Sharma et al.~\cite{sharma-2017}, a generative model for reconstructing 2D and 3D shapes. The input to the model is a voxelized representation of a point cloud. Inspired by the way human designers decompose objects, CSGNet aims to represent the underlying shape as a sequence of parametric primitives, such as circles or rectangles in 2D, and cuboids or cylinders in 3D, combined with operations including extrusion and deformation. Describing shapes using a small number of such primitives yields a compact and human-readable representation, which also facilitates editing and modification.  The model architecture comprises a CNN that encodes the voxelized input, followed by a Recurrent Neural Network (RNN) with Gated Recurrent Units (GRUs) to generate the sequence corresponding to the CSG tree. Despite its contributions, the model has several limitations. It operates on voxelized inputs rather than raw point clouds, which introduces discretization artifacts. Furthermore, the expressiveness of the generated shapes is limited by the fixed and finite set of predefined primitives. Finally, although pioneering, CSG is not the standard representation employed in most industrial CAD systems~\cite{khan-2024}. An approach to primitive fitting is presented in the Supervised Primitive Fitting Network (SPFN) by Li et al.~\cite{li-2018}, which aims to fit a set of geometric primitives to 3D point cloud data. This approach enriches otherwise unstructured point clouds with structural information by representing shapes using interpretable and editable primitives such as planes, spheres, cylinders, and cones. SPFN predicts, for each point in the input cloud, its corresponding normal vector as well as a primitive type label. These predictions are then used to regress the parameters of the underlying geometric primitives. The model employs PointNet++ as a backbone for feature extraction, supporting both the classification and regression tasks. However, the output consists of continuous primitive surfaces, requiring an additional triangulation step to obtain a mesh representation.

\begin{figure}[b]
    \centering
    \includegraphics[width=1\linewidth]{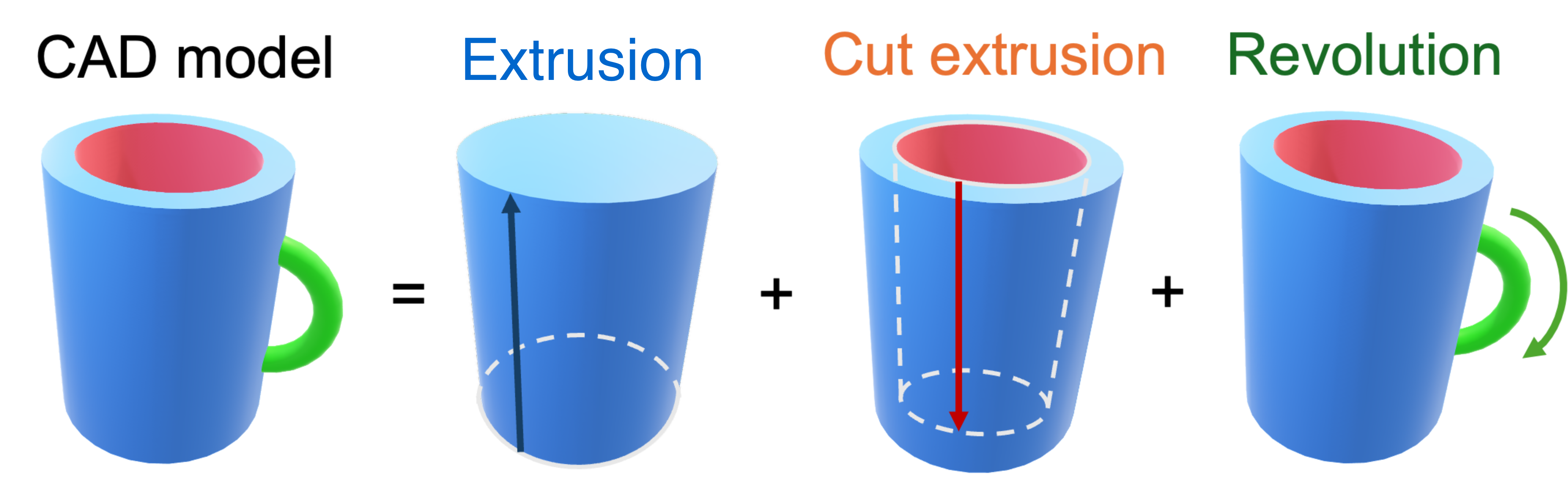}
    \caption{CAD modeling is a feature-based process}
    \label{fig:cad_process}
\end{figure}

In the area of CAD sequence reconstruction, DeepCAD by Wu et al.~\cite{Wu-2021_ICCV} represents pioneering work. The authors introduce a generative model capable of autoencoding CAD modeling sequences and propose a domain-specific language for describing such sequences. In addition, they release a large-scale dataset containing approximately 178{,}000 CAD samples. In their formulation, a CAD model is constructed as a sequence of sketch-and-extrude operations, closely reflecting typical user interactions with CAD software. In Figure~\ref{fig:cad_process} the feature-wise CAD modeling process is depicted. Each operation is defined by a command (e.g., sketch or extrude) and a set of associated arguments.  The model adopts a transformer-based autoencoder architecture, in which an encoder maps the CAD sequence into a latent space and a decoder reconstructs the sequence from this representation. Both the proposed language and the released dataset have become foundational resources for subsequent research in this field. In their discussion of future work, the authors suggest integrating a point cloud encoder to map raw 3D input into the same latent space, enabling the direct generation of CAD models from geometric data. The dataset and model architecture proposed in DeepCAD form an important foundation for the present work. 

Recently, Large Language Models (LLMs) have attracted significant attention due to their impressive capabilities in text and image generation. Consequently, researchers have begun exploring their application in the domain of CAD. In this context, many works focus on employing LLMs for CAD code generation, including CAD-specific notations as well as Python, SQL, C++, and Java code~\cite{zhang-2025}.  A recent example of LLM-based Python modeling is CAD-Recode~\cite{rukhovich-2025}, which extends the DeepCAD language into executable Python scripts that represent CAD modeling workflows. By leveraging pre-trained LLMs, CAD-Recode can infer Python-based CAD sequences, providing a novel interface between natural language, procedural code, and 3D design. In a related study, Li et al.~\cite{li-2024} provide images, sketches, and textual descriptions of CAD models as input to a Generative Pre-trained Transformer (GPT) model, specifically GPT-4. The authors instruct the model to act as an artificial intelligence assistant specialized in designing 3D objects using CadQuery, a Python-based CAD modeling language. In addition, they develop a parsing script to generate new samples suitable for LLM training in the context of CAD models.

Point2Cyl by Uy et al.~\cite{uy-2021} proposes a supervised network that decomposes a raw point cloud into a set of extrusion cylinders, following the sketch-and-extrude paradigm of CAD modeling. The model predicts per-point extrusion instance segmentation, base/barrel membership, and surface normals simultaneously. Using PointNet++ as an encoder, the extracted per-point features are processed by two parallel fully connected branches: one predicts geometric parameters (extrusion axis, center, sketch scale, and extrusion extent), and the other maps the input into a latent sketch space representing the 2D profile to be extruded. From these outputs, the extrusion parameters are recovered in a closed-form, differentiable formulation. Reconstruction is then performed either through a volumetric composition of the predicted sketch implicits or by loading the extracted primitives into a CAD modeler such as Fusion360 for further assembly and editing. A key limitation of this approach is that the sign of each extrusion cylinder, that is, whether it contributes additive or subtractive volume, cannot be determined automatically and must be resolved in post-processing. Furthermore, the method operates entirely in geometric space, is tailored to a specific network architecture, and does not produce structured CAD sequences that could be leveraged by or transferred to other reconstruction models.

The work of Uy et al. is the most closely related to the extrusion segmentation strategy presented in this paper. However, our approach differs fundamentally in both representation and methodology. We build on the CAD sequence representation introduced by Wu et al.~\cite{Wu-2021_ICCV}, in which a CAD model is expressed as a sequence of sketch-and-extrude commands, and we propose a novel sequence-level extrusion segmentation strategy as our key contribution. Rather than segmenting a point cloud into geometric regions such as base and barrel areas as done by Uy et al., we split each CAD sequence at every extrusion command, yielding a set of single-extrusion subsequences, referred to as primitive models, each paired with a corresponding point cloud sampled from the partial geometry. This increases training set diversity without requiring the construction of new datasets, and makes the learning task more challenging, as the input point clouds of primitive models are inherently incomplete, forcing the model to infer missing geometry and generate complete command sequences from partial observations. As a result, the model learns richer and more robust point cloud representations. At inference, the model processes each primitive point cloud independently and the final CAD sequence is obtained by concatenating the individual predictions, making assembly fully automatic. Crucially, our strategy is model-agnostic: it can be applied to any architecture that reconstructs CAD sequences from point clouds, offering a simple yet effective means of boosting reconstruction performance across a broad range of models and objects.

\section{CAD Reconstruction from Point Clouds}
\label{sec:methodology}
As a baseline for evaluating our extrusion segmentation strategy, we first reconstruct CAD models from point clouds without applying the proposed extrusion-based approach. To this end, we describe the sketch-and-extrude CAD representation, the reconstruction model, and the evaluation metrics used to assess performance.

\begin{table}[b!]
\centering
\small
\setlength{\tabcolsep}{4pt}
\renewcommand{\arraystretch}{1.15}
\begin{tabularx}{\columnwidth}{@{}l l X l@{}}
\toprule
\textbf{Command}
& \multicolumn{2}{c}{\textbf{Parameter description}}
& \textbf{Type} \\
\cmidrule(lr){2-3}


L
& $x, y$ & line end point
& numeric \\

\addlinespace
\multirow{3}{*}{A}
& $x, y$ & arc end point
& numeric \\
& $a$ & sweep angle
& numeric \\
& $f$ & counterclockwise flag
& boolean \\

\addlinespace
\multirow{2}{*}{R}
& $x, y$ & center
& numeric \\
& $r$ & radius
& numeric \\

\addlinespace
\multirow{6}{*}{E}
& $o, d, y$ & sketch plane orientation
& numeric \\
& $p_x, p_y, p_z$ & sketch plane origin
& numeric \\
& $s$ & sketch scale
& numeric \\
& $e_1, e_2$ & extrusion distances
& numeric \\
& $b$ & CSG operation
& boolean \\
& $u$ & extrusion type
& boolean \\


\bottomrule
\end{tabularx}
\caption{Commands and parameters}
\label{tab:commands_and_params}
\end{table}

\subsection{CAD Sequence}
To understand our approach to reconstructing CAD models from point clouds, as well as the proposed extrusion-based segmentation strategy, it is first necessary to clarify how CAD models can be expressed in a neural-network-friendly format. This representation, originally introduced by Wu et al.~\cite{Wu-2021_ICCV} and adopted in this work, reflects the sequential and feature-driven nature of user-level CAD modeling. In this formulation, a CAD model is represented as a sequence of sketch-and-extrude commands. Each sequence consists of up to six sketch-and-extrude operations, where each command comprises between 2 and 11 arguments selected from a total of 16 possible options, as summarized in Table~\ref{tab:commands_and_params}. Every extrusion sequence begins with a start-of-line (SOL) token, followed by one or more sketch commands (\textbf{L}ine, \textbf{A}rc, or Ci\textbf{R}cle), and concludes with an \textbf{E}xtrude command. The entire CAD sequence is terminated by an end-of-sequence (EOS) token. All numeric parameters, including coordinates and radii, are normalized and quantized into 256 discrete values, rendering them unitless. The sketch commands define closed profiles and specify a 2D sketch on a sketch plane. The Extrude command then converts this 2D sketch into a 3D shape by defining the orientation and origin of the sketch plane in 3D space, as well as the extrusion parameters, such as extrusion distance, extrusion type (one-, two-sided or symmetric), and the associated Constructive Solid Geometry (CSG) operation. The CSG operation determines how the newly created geometry interacts with existing bodies, including operations such as creating a new body, joining, cutting, or intersecting. Sketches encode the real-world proportions of the CAD model. Since CAD models may vary in scale, a normalization step is required to obtain a unified representation. For this purpose, all sketches are normalized to the unit square, and a predicted scaling factor, referred to as the sketch size $s$, is used to denormalize the sketches after inference. As a result, the model predicts CAD geometries within a unit cube without recovering their absolute dimensions, while preserving the relative proportions of the original shapes. In this work, a CAD sequence may contain up to 60 commands with a maximum of 10 extrusion operations. In Figure~\ref{fig:seg_strat} an example CAD sequence is shown. These sequences can be converted into explicit CAD models using the pythonOCC library~\cite{paviot2022pythonocc}.

\subsection{Reconstruction Model}

Our end-to-end baseline model for CAD reconstruction from point clouds combines PointNet++~\cite{Qi-2017_NIPS} with the DeepCAD decoder~\cite{Wu-2021_ICCV}. PointNet++ consists of multiple set abstraction layers that progressively refine features by selecting centroids using Farthest Point Sampling (FPS) and grouping local neighborhoods within a fixed radius. The points in each neighborhood are processed by Multi-Layer Perceptrons (MLPs) to expand the feature representation, while deeper layers further increase dimensionality. A global max-pooling operation aggregates the features into a permutation-invariant 256-dimensional latent representation of the point cloud. This latent vector is processed by a linear layer serving as a bottleneck and is then passed to the transformer-based DeepCAD decoder, which consists of four attention layers~\cite{Vaswani-2017_NIPS}. Positional embeddings for the 60 command positions are first added to the latent representation from PointNet++, and the resulting combined representation is sequentially refined by the individual attention layers. Finally, two parallel fully connected layers take the output of the attention layers and produce the logits for the command sequence and their corresponding parameters. Unlike the two-stage training of Wu~\cite{Wu-2021_ICCV}, our unified architecture removes the intermediate regression step.

\begin{figure}[b]
    \centering
    \includegraphics[width=1\linewidth]{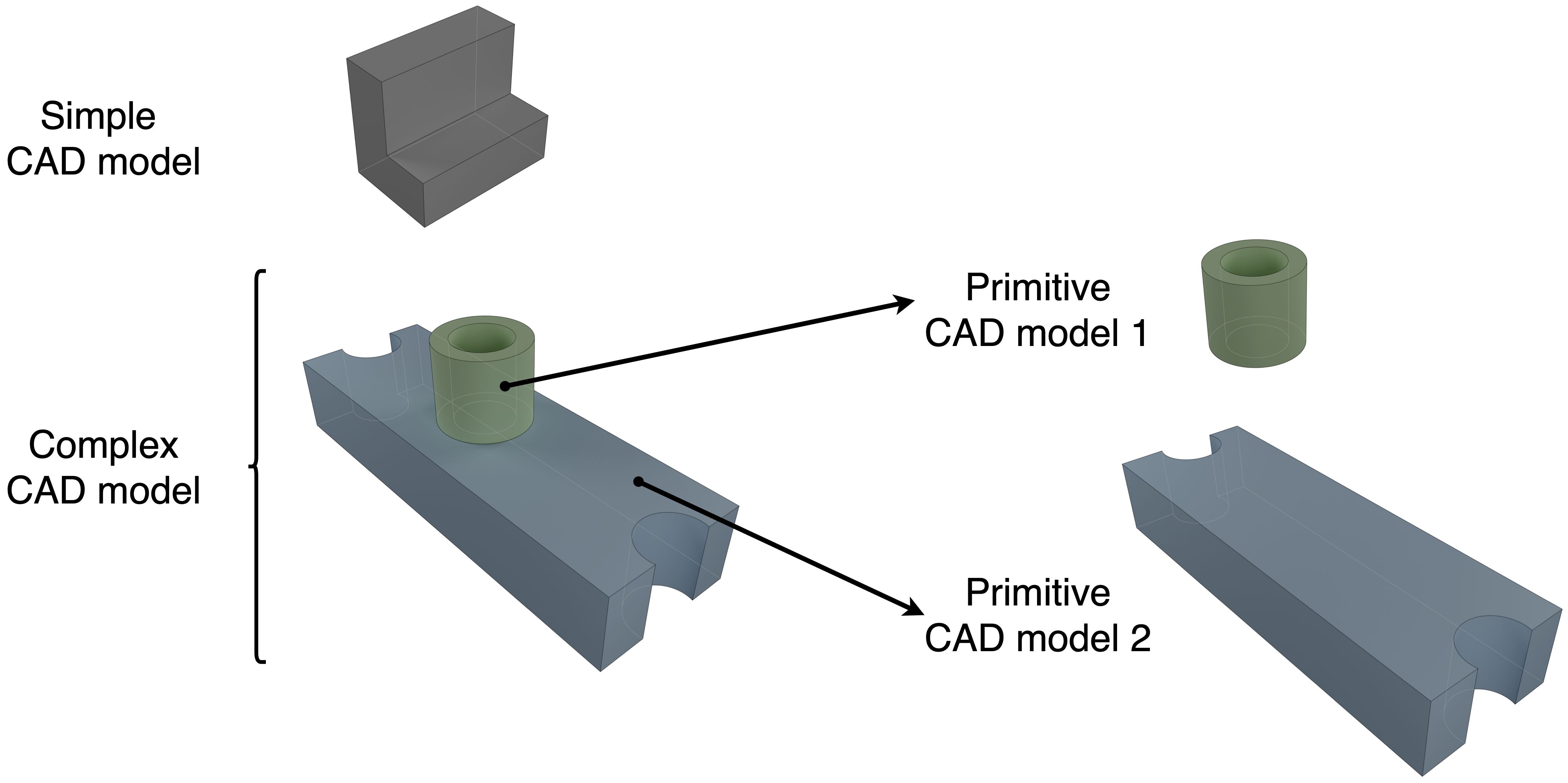}
    \caption{Difference between complex, simple and primitive CAD models.}
    \label{fig:simple_complex_primitive}
\end{figure}
\subsection{Evaluation Metrics}
The reconstruction quality is evaluated using command accuracy, parameter accuracy, the median Chamfer Distance (CD), and the invalid ratio (IR). Command and parameter accuracy measure how closely the predicted command and parameter sequences match the ground truth (GT). For command accuracy, the predicted command sequence is compared to the GT sequence up to the target length, and the number of correctly predicted commands is divided by the total number of target commands:
\begin{equation}
\label{eq:ACC_cmd}
    \text{ACC}_{\text{cmd}} = \frac{1}{N_c}\sum_{i = 1}^{N_c} \mathbb{I}\left[t_i = \hat{t}_i\right]
\end{equation}
where $N_c$ is the total number of target commands, $t_i$ is the $i$-th GT command type, $\hat{t}_i$ is the corresponding predicted command type, and $\mathbb{I}$ is the indicator function. Parameter accuracy is computed similarly, but only for parameters from correctly predicted commands. A predicted parameter is counted as correct if it lies within $\pm\eta = 3$ of the target value in quantized form, and the number of correct parameters is divided by the total number of parameters across all correctly predicted commands $K$, where $\lvert \mathbf{p}_i \rvert$ is the number of parameters of command $i$:
\[
K = \sum_{i=1}^{N_c} \mathbb{I}\left[t_i = \hat{t}_i\right] \cdot \lvert \mathbf{p}_i \rvert
\]
\begin{equation}
\label{eq:ACC_param}
    \text{ACC}_{\text{param}} = \frac{1}{K}\sum_{i = 1}^{N_c}\sum_{j = 1}^{\lvert \hat{p}_i\rvert} \mathbb{I}\left[\lvert p_{i,j} - \hat{p}_{i,j} \rvert < \eta \right] \cdot \mathbb{I}\left[t_i = \hat{t}_i\right]
\end{equation}
where $p_{i,j}$ and $\hat{p}_{i,j}$ denote the $j$-th GT and predicted parameter of the $i$-th command, respectively.
While command and parameter accuracy assess sequence-level correctness, the CD focuses on geometric fidelity. A single wrongly predicted command may reduce command accuracy only slightly yet alter the resulting shape significantly. The CD computes the average nearest-neighbor distance between the input point cloud and a point cloud sampled from the reconstructed model, where 2000 points are uniformly sampled from the surface of the predicted CAD model after exporting it to STEP format. The CD between the two point clouds \(P = \{p_i \in \mathbb{R}^3\}_{i=1}^{n}\) and \(Q = \{q_j \in \mathbb{R}^3\}_{j=1}^{m}\) is defined as: 
\begin{equation}
\label{eq:cd}
    \text{CD}(P,Q)=\frac{1}{n}\sum_{i=1}^{n}\left(\min_{q\in Q}\lvert\lvert p_i- q\rvert\rvert\right)^2+\frac{1}{m}\sum_{j=1}^{m}\left(\min_{p\in P}\lvert\lvert q_j-p\rvert\rvert\right)^2
\end{equation}
Finally, not all predicted CAD sequences are geometrically valid. Common sources of invalidity include a zero extrusion distance, a sketch of zero size, a degenerate profile with zero area (e.g., coincident opposing edges), or a cut extrusion that does not intersect any existing geometry. The IR quantifies this by measuring the proportion of predicted sequences that are invalid:
\begin{equation}
\label{eq:IR}
    \text{IR} = \frac{\text{Number of valid CAD-sequences}}{\text{Number of invalid CAD-sequences}}
\end{equation}
In conclusion, each metric captures a distinct aspect of reconstruction quality, but for a holistic evaluation, visual inspection of the results remains essential.

\section{Extrusion Segmentation Strategy}
\label{sec:segmentation}
Training our baseline model on the DeepCAD dataset yields high-fidelity reconstructions of non-complex objects, especially primitive shapes such as cubes and hollow or solid cylinders, as well as assemblies of these forms, which is shown in the results section. Yet reconstruction quality degrades on more complex shapes, which tend to integrate multiple features and exhibit more intricate geometric structures. To address this, we introduce a segmentation strategy that increases training-set diversity by splitting CAD sequences into their individual extrusion sequences and using these partial shapes to improve generalization. For this, we introduce the notation of complex, simple and primitive CAD models, shown in Figure~\ref{fig:simple_complex_primitive}: A CAD model is considered a complex model if more than one extrusion is necessary to create it. In contrast, each extrusion of a complex model is considered a primitive CAD model. However, if a CAD model can be created by only one extrusion, it is referred to as a simple CAD model. Thus, a simple model is a complete CAD model, whereas a primitive model is only a part of a complex one.

\begin{figure}[t]
    \centering
    \includegraphics[width=1.0\linewidth]{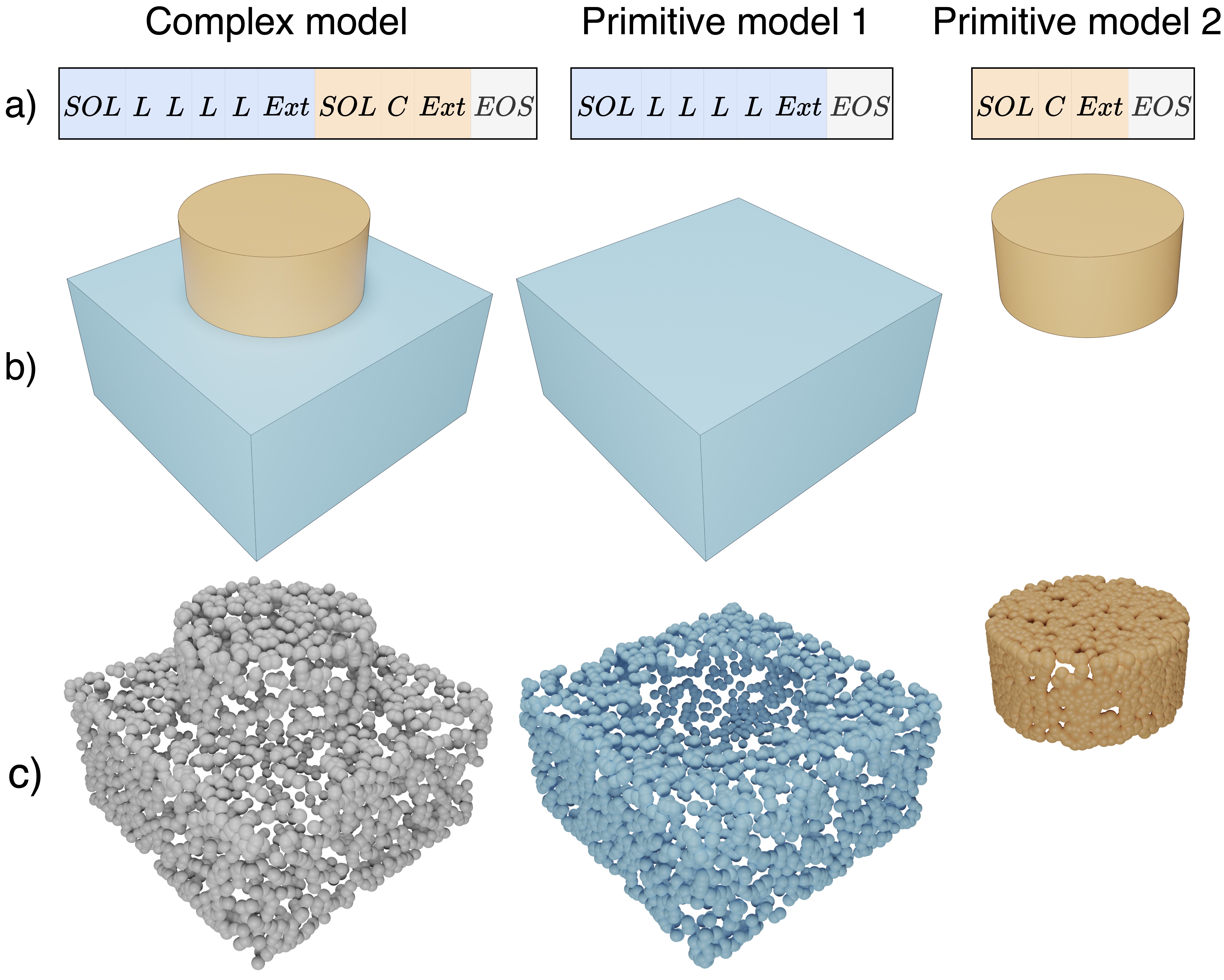}
    \caption{(a) CAD sequences (b) GT CAD models. (c) Input point clouds. For the primitive models, no points are present at their original interface. All point clouds of simple, complex and primitive models contain the same number of points.
}
    \label{fig:seg_strat}
\end{figure}

To train our end-to-end model, input point clouds must first be generated from the target CAD sequences. Using the pythonOCC library~\cite{paviot2022pythonocc}, CAD sequences are converted into 3D models from which uniform point clouds are sampled. For both simple and complex CAD models, this process is largely straightforward. However, for primitive CAD models, point sampling often produces invalid points, typically at interfaces between geometries or in regions created by cut extrusions. For example, a cube with a protruding cylinder is considered a complex CAD model because it results from two extrusions. When split into separate cube and cylinder primitives, sampling introduces unphysical points: on the cube surface where the cylinder would attach and, on the cylinder’s bottom face. These regions are not exposed in the complex model and should therefore not be sampled. The problem becomes even more pronounced with cut extrusions, which remove material defined by the intersection between the cut geometry and the existing model. If the cut is modeled independently, point sampling again produces unphysical points, since only newly exposed surfaces should contribute to the final point cloud. In Figure~\ref{fig:seg_strat} the process of splitting complex CAD models into primitive ones, alongside their point clouds is depicted

To address these issues, primitive model point clouds are generated by first densely sampling the point cloud of a complex CAD model. Each primitive, including cut extrusions, is then modeled individually and densely sampled. The resulting primitive point clouds are merged and overlaid with the densely sampled point cloud of the complete model. A nearest-neighbor search is subsequently performed, retaining only those points in each primitive point cloud that have neighbors within a specified radius (in this work $r=0.005$ was used) in the point cloud of the complex model. This procedure ensures that each primitive point cloud covers only surfaces that are present in the original CAD model, as illustrated in Figure~\ref{fig:seg_strat}c). Finally, each primitive point cloud is downsampled to 2{,}048 points while preserving its spatial position within the complex model. This allows the network to learn the correct placement of primitives during inference. Using this strategy, in addition to the 106{,}971 simple models provided in DeepCAD, 70{,}977 complex models yield a total of 255{,}281 primitive models. These primitive models significantly expand the training set and introduce greater variability, particularly because their geometry does not correspond one-to-one with the final CAD model. This makes the learning task less trivial and improves robustness.

In summary, a baseline multi-extrusion model (MEM) is trained on point clouds of both simple and complex CAD models. Our single-extrusion model (SEM), which uses the same architecture, is trained exclusively on point clouds of simple and primitive CAD models, that is, models containing only a single extrusion. To evaluate performance on complex models, the SEM is provided with the sequence of primitive point clouds of a complex model and processes each 2{,}048-point primitive sequentially. The final CAD sequence is then obtained by concatenating the individual predictions.

\section{Experimental Setup}
\label{sec:exp_setup}

The dataset used in this work is the DeepCAD dataset~\cite{Wu-2021_ICCV}, which comprises 178{,}238 CAD models stored in JavaScript Object Notation (JSON) format. Each JSON file encodes a CAD model as a sequence of commands and parameters, which are parsed and converted into OpenCascade Technology (OCCT) representations using the pythonOCC library~\cite{paviot2022pythonocc}. From these representations, explicit 3D models are constructed, and uniform point clouds are subsequently sampled with no noisde using the trimesh library~\cite{trimesh}. During this conversion process, 290 JSON files could not be successfully transformed due to faulty CAD models. Consequently, the final dataset used for the baseline and subsequent experiments contains slightly fewer samples. The dataset split follows the protocol of Wu et al.~\cite{Wu-2021_ICCV}, dividing the data into training, validation, and test sets with a ratio of 90\%--5\%--5\%, as summarized in Table~\ref{tab:baseline_dataset}.

Our extrusion segmentation strategy results in the Extrusion Segmentation dataset, in which each extrusion from the original DeepCAD dataset is treated as an independent CAD sequence paired with its corresponding point cloud. As described in Section~\ref{sec:segmentation}, primitive point clouds are generated by overlaying a densely sampled point cloud of the complete model and retaining only those points from each primitive's point cloud that fall within a fixed radius $r$ of a point on the complete model. A consequence of this procedure is that the number of points retained for a given primitive is not fixed, but is proportional to the fraction of the complete model's surface that the primitive occupies. For complex models containing one large and one small extrusion, both primitives are initially sampled with the same density. However, since the complete model's uniform point cloud naturally allocates fewer points to smaller surface regions, fewer points from the small primitive survive the radius-based filtering step. In extreme cases, where the complete model's point cloud contains no points in the vicinity of a small primitive, that primitive retains no valid points at all and cannot be paired with a point cloud. This results in 1{,}015 samples that could not be successfully generated. Nevertheless, this downsampling is necessary to ensure a fair comparison with the baseline approach, which also operates on a fixed number of points per point cloud, and given the overall dataset size of more than 300{,}000 samples, the impact of these missing samples is considered negligible.
\sisetup{
    group-separator = {,},
    group-minimum-digits = 4 
}

\begin{table}[t]
\centering
\small
\setlength{\tabcolsep}{6pt}
\begin{tabular}{
    l
    S[table-format=6,group-separator={,}, table-column-width=1cm]
    S[table-format=5,group-separator={,}, table-column-width=1.2cm]
    S[table-format=5,group-separator={,}, table-column-width=1cm]
}
\toprule
\textbf{Dataset Version} & \textbf{Training} & \textbf{Validation} & \textbf{Testing} \\
\midrule
Original DeepCAD        & 160982 & 8928  & 8038  \\
Extrusion Segmentation  & 327393 & 18236 & 16596 \\
\bottomrule
\end{tabular}
\caption{Number of samples in each dataset. Using the segmentation strategy, the dataset size is increased, without creating new data.}
\label{tab:baseline_dataset}
\end{table}

\begin{figure}[h]
    \centering
    \includegraphics[width=1\linewidth]{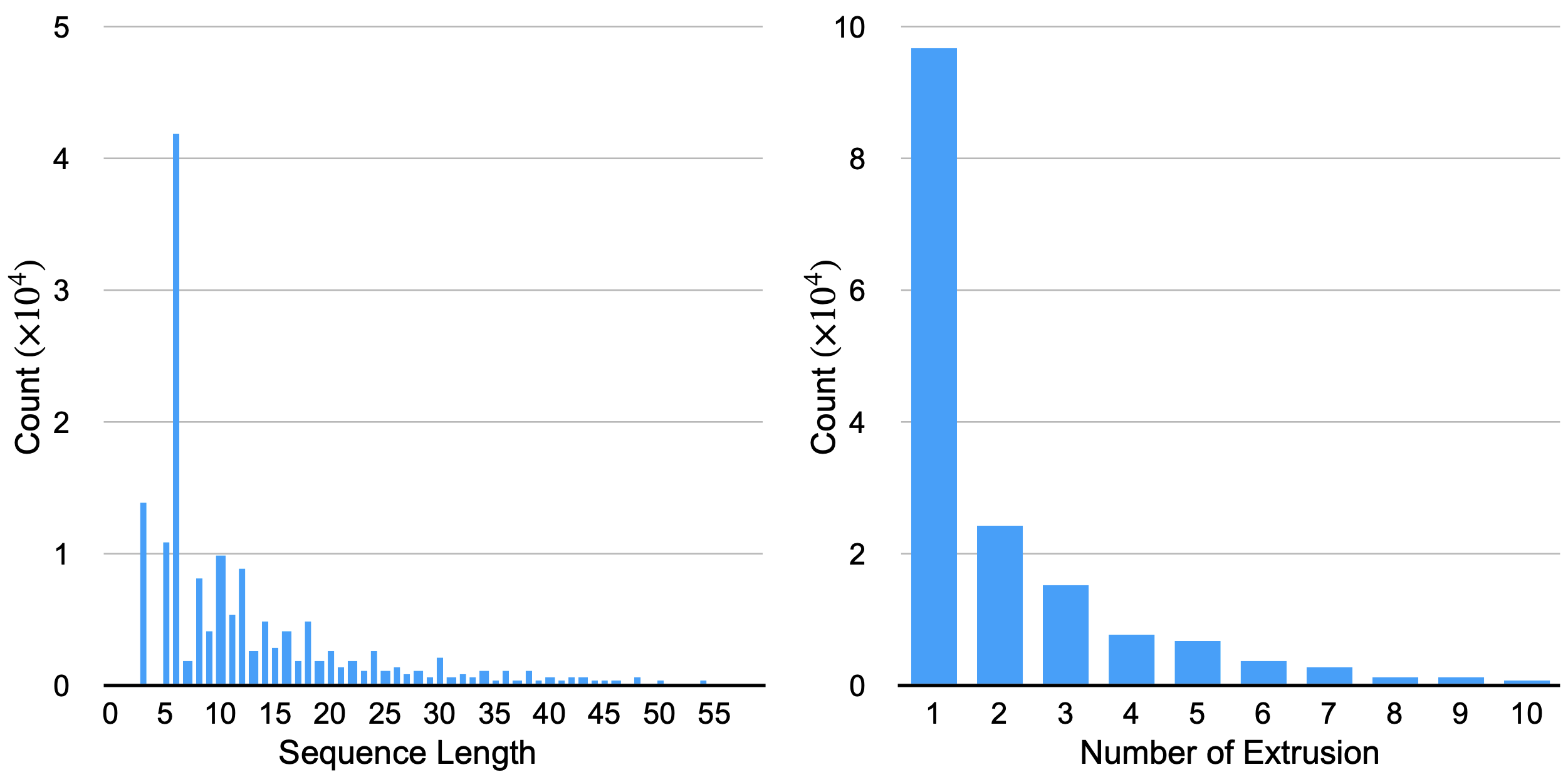}
    \caption{Number of samples per sequence length and per number of extrusions in the original DeepCAD dataset. Image courtesy~\protect\cite{Wu-2021_ICCV}}
    \label{fig:deepcad_ed}
\end{figure}

At the time of its release, the DeepCAD dataset was, alongside the Fusion360 Gallery Reconstruction dataset~\cite{Fusion}, which contains significantly fewer samples (8{,}625), one of the only datasets providing CAD models together with their modeling history. This made it a foundational resource for subsequent research in CAD reconstruction. However, the dataset is unbalanced and presents several inherent challenges. Figure~\ref{fig:deepcad_ed} shows histograms of the number of samples by sequence length and number of extrusions. The dataset is heavily biased toward simple CAD models with sequence lengths between 3 and 15, which can hinder the DL model's ability to learn effective representations for more complex, multi-extrusion CAD models. A thorough visual inspection of numerous samples reveals that many models resemble basic geometric primitives such as cubes, cuboids, solid and hollow cylinders, or other types of prisms. Many multi-extrusion models are composed of simple shapes that are stacked, intersected, or subtracted from each other. Some models exhibit more abstract forms, while others resemble real-world objects such as tables, chairs, wrenches, stick figures, or cameras. Additionally, several samples resemble components commonly found in manufacturing, including plates with stamped holes or bent metal parts. However, a number of models also display quality issues. Some contain disconnected parts, where one component appears to float above another. Others exhibit modeling errors, such as non-intersecting cut extrusions or extremely thin features that are likely to be lost during quantization.

To implement the model architecture described above, the DeepCAD bottleneck and decoder by Wu et al.~\cite{Wu-2021_ICCV} were used without modification. For the PointNet++ model, a third-party PyTorch implementation available on GitHub~\cite{PN_Github} was adapted for the tasks in this work using the multi-stage-grouping (MSG) configuration. This configuration follows a simplified MSG design, as suggested by Wu et al.~\cite{Wu-2021_ICCV}, employing only one radius per Set Abstraction (SA) layer. Empirical tests indicated that more complex variants resulted in decreased performance. Additionally, a \texttt{tanh} activation function is applied to the output of the final SA layer. The PointNet++ model consists of three SA layers configured as follows:

\begin{itemize}
    \item \textbf{SA1:} Centroids = 512, Radius = 0.1, Neighbor Points = 64, MLP = 32-32-64
    \item \textbf{SA2:} Centroids = 256, Radius = 0.2, Neighbor Points = 64, MLP = 128-128-256
    \item \textbf{SA3:} Centroids = 128, Radius = 0.4, Neighbor Points = 64, MLP = 128-128-256
\end{itemize}

All models are implemented in Python using the PyTorch framework~\cite{pytorch} and trained with CUDA~\cite{cuda} on the Phoenix Cluster\footnote{\url{https://www.tu-braunschweig.de/it/hpc}} at TU Braunschweig, utilizing four Tesla P100-SXM2-16GB GPUs. All models are trained using cross-entropy loss and optimized with the Adam optimizer~\cite{kingma-2014}. The number of trainable parameters for each model is summarized in Table~\ref{tab:model_params}.

\begin{table}[t]
\centering
\small
\setlength{\tabcolsep}{6pt}
\sisetup{
    group-separator = {,},
    detect-all 
}
\begin{tabular}{l l S[table-format=7.0, table-number-alignment = center]}
\toprule
\textbf{Module} & \textbf{Component} & \textbf{Trainable Parameters} \\
\midrule
DeepCAD     & Bottleneck & 65792 \\
DeepCAD     & Decoder    & 3445782 \\
\midrule
PointNet++  & SA1-SA3    & 1534624 \\
\bottomrule
\end{tabular}
\caption{Number of trainable parameters per module/component.}
\label{tab:model_params}
\end{table}

In summary, the baseline multi-extrusion model (MEM) is trained on point clouds of both simple and complex CAD models. Our single-extrusion model (SEM), which shares the same architecture, is trained exclusively on point clouds of simple and primitive CAD models to evaluate the impact of the proposed segmentation strategy. During evaluation on complex models, the SEM processes each 2{,}048-point primitive sequentially, and the final CAD sequence is obtained by concatenating the individual predictions.  Both models are trained using a learning rate scheduler, early stopping, and a mechanism to save the best-performing model. The batch size is set to 48 per GPU. A weighted average of validation command accuracy and argument accuracy is used as the monitoring metric, with equal weights of $0.5$, ensuring that command and argument performance are treated equally. The initial learning rate is \(1 \times 10^{-4}\) and is reduced by a factor of 0.1 after 30 epochs without improvement in the monitoring metric. This composite metric is employed for early stopping (after 50 epochs), learning rate scheduling, and best model checkpointing. The models required different training durations: the MEM converges in 44 hours, while the SEM requires 263 hours.

\section{Results}
\label{sec:results}
The results in Table~\ref{tab:results_comp} show that the SEM outperforms the MEM on both simple and complex CAD models across all metrics. Improvements are most pronounced for complex models, with command accuracy increasing by nearly 20 points, parameter accuracy by 12.56 points, IR reduced by more than half, and a notably lower median CD. In particular, the SEM exhibits a substantially lower CD and more stable performance as sequence length increases, indicating stronger generalization to complex shapes. Qualitative results confirm that this model better captures fine geometric details and more accurately places individual primitives, especially when complex objects are composed of multiple simple extrusions. In contrast, the multi-extrusion model, although benefiting from holistic spatial context, frequently oversimplifies intricate geometries or misinterprets structural components, leading to missing or incorrectly reconstructed details. 

\begin{figure}[t!]
    \centering
    \includegraphics[width=1\linewidth]{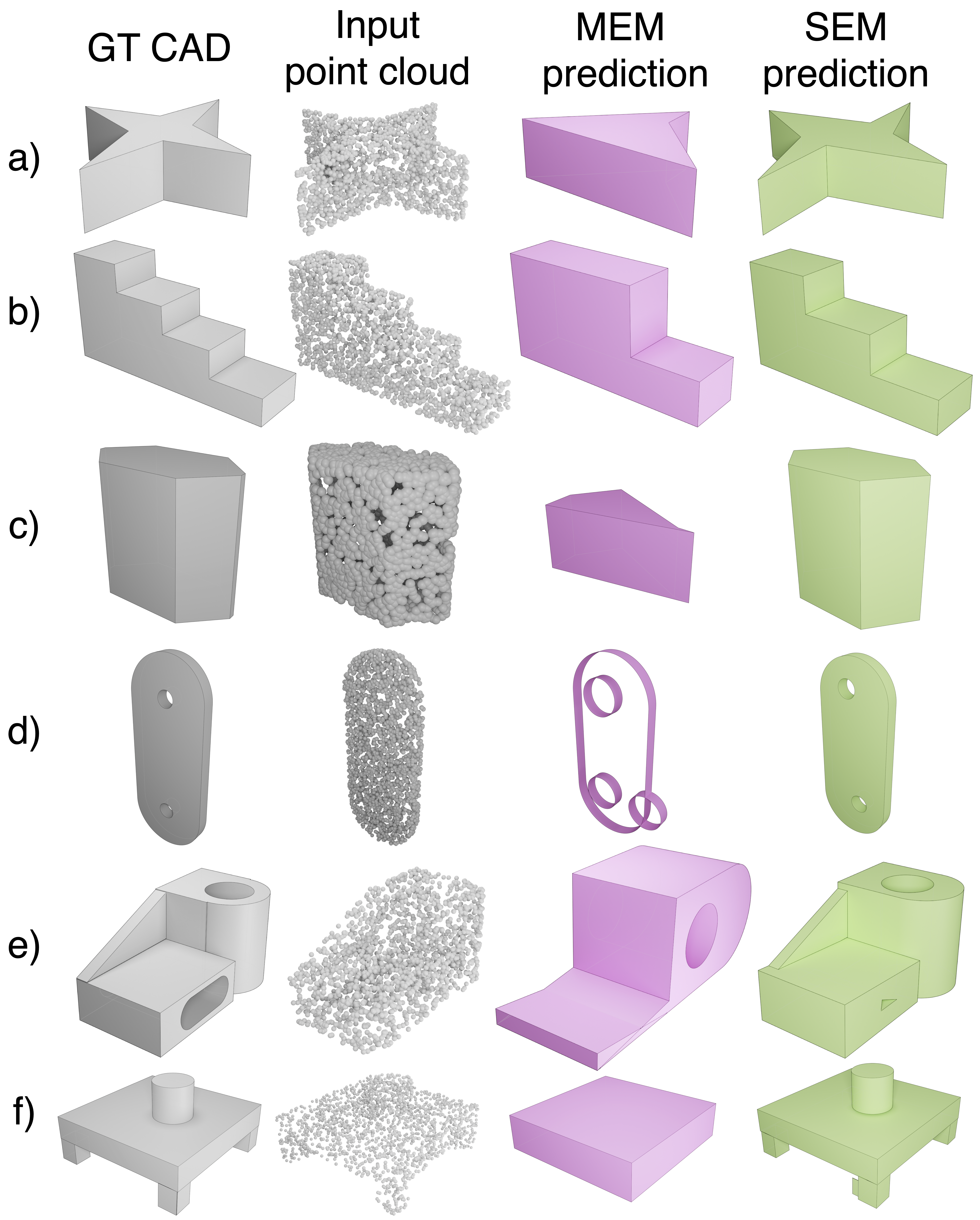}
    \caption{Comparison of GT, MEM output, and SEM output on (a, b, c, d) simple and (e, f) complex CAD models.}
    \label{fig:comparison}
\end{figure}

Performance differences for simple models are smaller but more meaningful, as both methods received identical inputs, highlighting the impact of our segmentation strategy. Here, command accuracy increases by 2.41 points and parameter accuracy by 5.13 points, while the median CD drops by 36\% and the IR is nearly halved, now 4 points lower. Reconstruction fidelity is highest for shorter command sequences and decreases as sequence length increases. Visual inspection shows that the model consistently captures finer geometric details, such as accurate distances, sharp outlines, and small features like holes, which are often missed by the other approaches. Since the single- and multi-extrusion models share the same architecture, the observed performance gap can be attributed to differences in training data, highlighting the advantage of training exclusively on primitive and simple CAD models. Nevertheless, arc-based commands remain a common source of error across all methods, and very small geometric features are frequently missed due to limited point cloud resolution. Visual results in Figure~\ref{fig:comparison} further confirm that the SEM captures point cloud features more effectively, benefiting from the additional primitive models in the training data and yielding higher-fidelity reconstructions. The code to obtain these results is available on GitHub: \href{https://github.com/saidharb/Point-Cloud-Reconstruction.git}{https://github.com/saidharb/Point-Cloud-Reconstruction.git}

\begin{table}[t]
\centering
\setlength{\tabcolsep}{3.5pt}

\begin{tabular}{@{} l
  S[table-format=2.2] S[table-format=2.2]
  S[table-format=2.2] S[table-format=2.2]
  S[table-format=2.2] S[table-format=2.2]
@{}}
\toprule
 & \multicolumn{2}{c}{\makecell{Simple\\CAD models}}
 & \multicolumn{2}{c}{\makecell{Complex\\CAD models}}
 & \multicolumn{2}{c}{\makecell{All\\CAD models}} \\
\cmidrule(rr){2-3} \cmidrule(rr){4-5} \cmidrule(rr){6-7}
 & \multicolumn{1}{c}{MEM}
 & \multicolumn{1}{c}{SEM}
 & \multicolumn{1}{c}{MEM}
 & \multicolumn{1}{c}{SEM}
 & \multicolumn{1}{c}{MEM}
 & \multicolumn{1}{c}{SEM} \\
\midrule
Cmd. acc. (\%)        & 88.03 & \textbf{90.44}  & 58.10 & \textbf{77.71} & 68.10 & \textbf{81.89}\\
Par. acc. (\%)       & 82.78 & \textbf{87.91}   & 69.94 & \textbf{82.50} & 75.47 & \textbf{84.25}\\
Med. CD $\times 10^{3}$& 1.54 & \textbf{0.98}     & 10.92 & \textbf{2.07} & 2.30 & \textbf{1.26} \\
IR (\%)               & 10.22 & \textbf{6.22}   & 32.02 & \textbf{12.03} & 20.22 & \textbf{8.66} \\
\bottomrule
\end{tabular}
\caption{Comparison of MEM and SEM performance on simple, complex, and all CAD models.}
\label{tab:results_comp}
\end{table}

\section{Conclusions}
\label{sec:conclusion}
Our segmentation strategy, to introduce primitive CAD models into the training set, improves the generalization of DL reconstruction models on unseen data. Despite not seeing complex models during training, the SEM outperformed the MEM on complex CAD models. For simple models, present in both training sets and without prior segmentation for the SEM, the improved metrics highlight the effectiveness of our approach.  Primitive CAD models are particularly valuable because their input point clouds can be incomplete, as surfaces that are not exposed in the original complex model are missing. This forces the model to infer missing regions and generate complete command sequences, promoting richer and more meaningful point cloud representations. By generating primitive CAD models and sampling their point clouds, training diversity is increased without needing to construct entirely new datasets. Since our approach operates purely on the data level and is independent of any specific model architecture, it can be applied to any DL model that reconstructs CAD sequences. Thus, our segmentation strategy provides a simple but effective way to leverage the full potential of CAD datasets for DL reconstruction models. 

As many CAD models come with their design history, our extrusion segmentation strategy can be employed in many contexts. It can be used for example in scan-to-BIM, infrastructure inspection, digital heritage preservation and reverse engineering of mechanical parts. Also in autonomous driving and robotics scenarios, improved reconstruction quality using our strategy can be of great advantage. In summary, the key advantage of our extrusion segmentation strategy is that it provides a simple yet effective means of improving deep learning model performance, and can be applied to any CAD dataset for which design history is available.

For future work, point cloud segmentation could be integrated into the SEM pipeline to enable reconstruction of complex CAD models without pre-segmented inputs, with our work supplying the necessary labels. Additionally, all input point clouds currently contain 2,048 points regardless of object size, allocating points proportionally to surface area could reduce bias toward smaller models. Finally, simulating laser-scanned point clouds from CAD models would enable training on data more representative of real-world scans, potentially contributing to models robust enough for real-world deployment.

\textbf{Acknowledgment}: This research is funded by the Deutsche Forschungsgemeinschaft (DFG, German Research Foundation) – TRR 277/2 2024 – Project number 414265976. The authors thank the DFG for the support within the CRC / Transregio 277 - Additive Manufacturing in Construction (Project C06: Integration of Additive Manufacturing into a Cyber-Physical Construction System).

{
	\begin{spacing}{1.17}
		\normalsize
		\bibliography{ISPRSguidelines_authors} 
	\end{spacing}
}

\end{document}